\documentclass[11pt]{article}
\usepackage[margin=1in]{geometry}

\usepackage{times}
\usepackage{url}
\usepackage{latexsym}
\usepackage{xcolor}
\usepackage{balance}
\usepackage{graphicx}
\usepackage{epstopdf}
\usepackage{amsmath}
\usepackage{color}
\usepackage{amsfonts}
\usepackage{amssymb}

\usepackage{enumitem}
\usepackage{mathtools}
\usepackage{multirow}
\usepackage{natbib}
\usepackage[font=small]{caption}

\newcommand{\denselist}{\setlength{\itemsep}{1pt}
  \setlength{\parskip}{0pt} \setlength{\parsep}{0pt}}
\newcommand{\bitem}{\begin{itemize}[noitemsep,topsep=2pt]\denselist}
\newcommand{\eitem}{\end{itemize}}

\title{SyntaxNet Models for the CoNLL 2017 Shared Task}

\author{
  Chris Alberti, 
  Daniel Andor, 
  Ivan Bogatyy, 
  Michael Collins,
  Dan Gillick, \\
  Lingpeng Kong\footnote{Carnegie Mellon University, Pittsburgh, PA.}, 
  Terry Koo, 
  Ji Ma, 
  Mark Omernick, 
  Slav Petrov, \\
  Chayut Thanapirom,
  Zora Tung, 
  David Weiss \\\\
  Google Inc\\
  New York, NY\\
}

\date{}

\begin{document}
\maketitle

\begin{abstract}
  We describe a baseline dependency parsing system for the CoNLL2017 Shared
  Task. This system, which we call ``ParseySaurus,'' uses the DRAGNN framework
  \citep{dragnn} to combine transition-based recurrent parsing and tagging with
  character-based word representations. On the v1.3 Universal Dependencies
  Treebanks, the new system outpeforms the publicly available, state-of-the-art
  ``Parsey's Cousins'' models by 3.47\% absolute Labeled Accuracy Score (LAS)
  across 52 treebanks.
\end{abstract}

\graphicspath{{./figs/}}
\newcommand{\bX}{\mathbf{X}}
\newcommand{\bE}{\mathbf{E}}
\newcommand{\bb}{\mathbf{b}}
\newcommand{\bH}{\mathbf{H}}
\newcommand{\bW}{\mathbf{W}}
\newcommand{\bh}{\mathbf{h}}
\newcommand{\mwords}{\mathrm{word}}
\newcommand{\mtags}{\mathrm{tag}}
\newcommand{\mlabels}{\mathrm{label}}
\newcommand{\todo}[1]{{\bf \color{red}{TODO: #1}}}
\newcommand{\eat}[1]{\ignorespaces}
\newcommand{\commentout}[1]{}
\newcommand\T{\rule{0pt}{4ex}}  

\section{Introduction}

Universal Dependencies\footnote{http://universaldependencies.org/} are growing in popularity
due to the cross-lingual consistency and large language coverage of the provided data.
The initiative has been able to connect researchers across the globe and now includes
64 treebanks in 45 languages.
It is therefore not surprising that the Conference on Computational Natural Language Learning (CoNLL) in 2017
will feature a shared task on ``Multilingual Parsing from Raw Text to Universal Dependencies.''

To facilitate further research on multilingual parsing and to enable even small
teams to participate in the shared task, we are releasing baseline
implementations corresponding to our best models.
This short paper describes (1) the model structure employed in these models, (2)
how the models were trained and (3) an empirical evaluation comparing these
models to those in \cite{andor2016globally}.
Our model uses on the DRAGNN framework \citep{dragnn} to improve upon
\cite{andor2016globally} with dynamically constructed, recurrent
transition-based models. The code as well as the pretrained models is available
at the SyntaxNet github repository.\footnote{
  https://github.com/tensorflow/models/tree/master/syntaxnet}.

We note that this paper describes the parsing model used in the
baseline. Further releases will describe any changes to the segmentation model
compared to SyntaxNet.

\section{Character-based representation}

Recent work has shown that learned sub-word representations can improve over
both static word embeddings and manually extracted feature sets for describing
word morphology.
\citet{jozefowicz2016exploring} use a convolutional model over the characters in
each word for language modeling.
Similarly, \citet{ling-EtAl:2015:EMNLP,ling2015character} use a bidirectional
LSTM over characters in each word for parsing and machine translation.

\citet{chung2016hierarchical} take a more general approach.
Instead of modeling each word explicitly, they allow the model to
learn a hierarchical ``multi-timescale'' representation of the input, where each
layer corresponds to a (learned) larger timescale.

Our modeling approach is inspired by this multi-timescale architecture in that
we generate our computation graph dynamically, but we define the timescales explicitly.
The input layer operates on characters and the subsequent layer operates on words,
where the word representation is simply the hidden state computed by the first layer
at each word boundary.
In principle, this structure permits fully dynamic word representations based on
left context (unlike previous work) and simplifies recurrent computation at the
word level (unlike previous work with standard stacked LSTMs
\citep{gillick2015multilingual}).

More related work
\citep{kim2015character,miyamoto2016gated,lankinen2016character,ling-EtAl:2015:EMNLP}
describes alternate neural network architectures for combining character- and
word-level modeling for various tasks. Like \cite{BallesterosDS15}, we use
character-based LSTMs to improve the Stack-LSTM \cite{dyer2015transition} model
for dependency parsing, but we share a single LSTM run over the entire sentence.

\section{Model}

Our model combines the recurrent multi-task parsing model of \citet{dragnn} with
character-based representations learned by a LSTM. Given a tokenized
text input, the model processes as follows:

\bitem
\item A single LSTM processes the entire character string (including
  whitespace)\footnote{In our UD v1.3 experiments, the raw text string is not
    available. Since we use gold segmentations, the whitespace is artificially
    induced, and functions as a ``new word'' signal for languages with no
    naturally occuring whitespace.}
  left-to-right. The last hidden state in a given token (as given by the word
  boundaries) is used to represent that word in subsequent parts of the model.
\item A single LSTM processes the word representations (from the first step) in
  right-to-left order. We call this the ``lookahead'' model.
\item A single LSTM processes the {\em lookahead} representations
  right-to-left. This LSTM has a softmax layer which is trained to predict POS
  tags, and we refer to it as the ``tagger'' model.
\item The recurrent compositional parsing model \citep{dragnn} predicts
  parse tree left-to-right using the {\em arc-standard} transition system. Given
  a stack $s$ and a input pointer to the buffer $i$, the parser dynamically
  links and concatenates the following input representations:
  \begin{itemize}
  \item Recurrently, the two steps that last modified the $s_o$ and $s_1$
    (either SHIFT or REDUCE operations). 
  \item From the {\em tagger} layer, the hidden representations for $s_0$, $s_1$, and $i$.
  \item From the {\em lookahead} layer, the hidden representation for $i$.
  \item All are projected to 64 dimensions before concatenating.
  \item The parser also extracts 12 discrete features for previously predicted
    parse labels, the same as in \cite{dragnn}.
  \end{itemize}
  At inference time, we use beam decoding in the parser with a beam size of
  8. We do not use local normalization, and instead train the models with
  ``self-normalization'' (see below).
\eitem

This model is implemented using the DRAGNN framework in TensorFlow. All code is
publicly available at the SyntaxNet repository. The code provides tools to
visualize the unrolled structure of the graph at run-time.

\subsection{Training}

We train using the multi-task, maximum-likelihood ``stack-propagation'' method
described in \citet{dragnn} and \citet{zhang2016stack}. Specifically, we use the
gold labels to alternate between two updates:

\begin{enumerate}
\item \textsc{Tagger}: We unroll the first three LSTMs and backpropagate
  gradients computed from the POS tags.
\item \textsc{Parser}: We unroll the entire model, and backpropagate gradients
  computed from the oracle parse sequence.
\end{enumerate}

We use the following schedule: pretrain \textsc{Tagger} for 10,000
iterations. Then alternate \textsc{Tagger} and \textsc{Parser} updates at a
ratio of 1:8 until convergence. 

To optimize for beam decoding, we regularize the softmax objective to be
``self-normalized.''\cite{vaswani2013decoding,andreas2015and}. With this
modification to the softmax, the log scores of the model are encouraged (but not
constrained) to sum to one. We find that this helps mitigate some of the bias
induced by local normalization \cite{andor2016globally}, while being fast and
efficient to train.

\subsection{Hyperparameters}

Like the ratio above, many hyperparameters, including design decisions, were
tuned to find reasonable values before training all 64 baseline models. While
the full recipe can be deciphered from the code, here are some key points for
practitioners:

\begin{itemize}
\item We use Layer Normalization \citep{ba2016layer} in all of our networks, both
  LSTM and the recurrent parser's Relu network cell.
\item We always project the LSTM hidden representations down from 256$\rightarrow$64 when we pass from one component to another.
\item We use moving averages of parameters at inference time.
\item We use the following ADAM recipe: $\beta_1 = \beta_2 = 0.9$, and set
  $\epsilon$ to be one of $10^{-3}, 10^{-4}, 10^{-5}$ (typically $10^{-4}$).
\item We normalize all gradients to have unit norm {\em before} applying the
  ADAM updates.
\item We use dropout both recurrently and on the inputs, at the same rate
  (typically 0.7 or 0.8).
\item We use a minibatch size of 4, with 4 asynchronous training threads doing
  asynchronous SGD.
\end{itemize}
\section{Comparison to Parsey's Cousins}

Since the test set is not available for the contest, we use v1.3 of the
Universal Dependencies treebanks to compare to prior state-of-the-art on 52
languages. Our results are in Table \ref{tab:ud}. We observe that the new
model outperforms the original SyntaxNet baselines, sometimes quite dramatically
(e.g. on Latvian, by close to 12\% absolute LAS.) We note that this is not an
exhaustive experiment, and further study is warranted in the future. Nonethelss,
these results show that the new baselines compare very favorably to at least one
publicly available state-of-the-art baseline.

\begin{table}
  \scalebox{0.9}{
  \begin{tabular}{ccccc|ccccc}
    \hline
    & \multicolumn{2}{c}{Parsey} & \multicolumn{2}{c}{ParseySaurus} & & \multicolumn{2}{c}{Parsey} & \multicolumn{2}{c}{ParseySaurus}\\
  Language & UAS & LAS & UAS & LAS & Language & UAS & LAS & UAS & LAS \\
    \hline 
    Ancient Greek-PROIEL	& 78.74	& 73.15	& 81.14	& 75.81	& Indonesian	& 80.03	& 72.99	& 82.55	& 76.31 \\
    Ancient Greek	& 68.98 & 62.07 & 73.85 & 68.1	& Irish	& 74.51	& 66.29	& 75.71	& 67.13 \\
    Arabic	& 81.49	& 75.82	& 85.01	& 79.8	& Italian	& 89.81	& 87.13	& 91.14	& 88.78 \\
    Basque	& 78.00	& 73.36	& 82.05	& 78.72	& Kazakh	& 58.09	& 43.95	& 65.93	& 52.98 \\
    Bulgarian	& 89.35	& 85.01	& 90.87	& 86.87	& Latin-ITTB	& 84.22	& 81.17	& 88.3	& 85.87 \\
    Catalan	& 90.47	& 87.64	& 91.87	& 89.7	& Latin-PROIEL	& 77.60	& 70.98	& 80.27	& 74.29 \\
    Chinese	& 76.71	& 71.24	& 81.04	& 76.56	& Latin	& 56.00	& 45.80	& 63.49	& 52.52 \\
    Croatian	& 80.65	& 74.06	& 82.84	& 76.78	& Latvian	& 58.92	& 51.47	& 69.96	& 63.29 \\
    Czech-CAC	& 87.28	& 83.44	& 89.26	& 85.26	& Norwegian	& 88.61	& 86.22	& 90.69	& 88.53 \\
    Czech-CLTT	& 77.34	& 73.40	& 79.9	& 75.79	& Old Church Slavonic	& 84.86	& 78.85	& 87.1	& 81.47 \\
    Czech	& 89.47	& 85.93	& 89.09	& 84.99	& Persian	& 84.42	& 80.28	& 86.66	& 82.84 \\
    Danish	& 79.84	& 76.34	& 81.93	& 78.69	& Polish	& 88.30	& 82.71	& 91.86	& 87.49 \\
    Dutch-LassySmall	& 81.63	& 78.08	& 84.02	& 80.53	& Portuguese-BR	& 87.91	& 85.44	& 90.52	& 88.55 \\
    Dutch	& 77.70	& 71.21	& 79.89	& 74.29	& Portuguese	& 85.12	& 81.28	& 88.33	& 85.07 \\
    English-LinES	& 81.50	& 77.37	& 83.74	& 80.13	& Romanian	& 83.64	& 75.36	& 87.41	& 79.89 \\
    English	& 84.79	& 80.38	& 87.86	& 84.45	& Russian-SynTagRus	& 91.68	& 87.44	& 92.67	& 88.68 \\
    Estonian	& 83.10	& 78.83	& 86.93	& 83.69	& Russian	& 81.75	& 77.71	& 84.27	& 80.65 \\
    Finnish-FTB	& 84.97 & 80.48	& 88.17	& 84.5 & Slovenian-SST	& 65.06	& 56.96	& 68.72	& 61.61 \\
    Finnish	& 83.65	& 79.60	& 86.96	& 83.96	& Slovenian	& 87.71	& 84.60	& 89.76	& 87.69 \\
    French	& 84.68	& 81.05	& 86.61	& 83.1 & Spanish-AnCora	& 89.26	& 86.50	& 91.06	& 88.89 \\
    Galician	& 84.48	& 81.35	& 86.22	& 83.65	& Spanish	& 85.06	& 81.53	& 87.16	& 84.03 \\
    German	& 79.73	& 74.07	& 84.12	& 79.05	& Swedish-LinES	& 81.38	& 77.21	& 83.97	& 79.93 \\
    Gothic	& 79.33	& 71.69	& 82.12	& 74.72	& Swedish	& 83.84	& 80.28	& 86.58	& 83.48 \\
    Greek	& 83.68	& 79.99	& 86.2	& 82.66	& Tamil	& 64.45	& 55.35	& 69.53	& 60.78 \\
    Hebrew	& 84.61	& 78.71	& 86.8	& 81.85	& Turkish	& 82.00	& 71.37	& 83.96	& 74.06 \\
    Hindi	& 93.04	& 89.32	& 93.82	& 90.23	& &&&& \\
    Hungarian	& 78.75	& 71.83	& 81.82	& 76.22	& {\bf Average}	& {\bf 81.12} & {\bf 75.85 }& {\bf 84.07 }& {\bf 79.33} \\
    \hline
  \end{tabular}
  }
  \caption{Comparison to prior SyntaxNet models on UD v1.3. On average, we observe a 14.4\% relative reduction in error (RRIE), 
   or 3.47\% absolute increase in LAS.}
  \label{tab:ud}
\end{table}

\subsection{CoNLL2017 Shared Task}

We provide pre-trained models for all 64 treebanks in the CoNLL2017 Shared Task
on the SyntaxNet website. All source code and data is publicly available. Please
see the task website for any updates.


\section*{Acknowledgements}

We thank our collaborators at the Universal Dependencies and Conll2017 Shared
task, as well as Milan Straka of UD-Pipe. We'd also like to thank all members of
the Google Parsing Team (current and former) who made this release possible.

\balance
\bibliographystyle{plainnat}
\bibliography{paper}

\end{document}